\documentclass[10pt,twocolumn,letterpaper]{article}

 \usepackage{cvpr}              
\definecolor{cvprblue}{rgb}{0.21,0.49,0.74}
\usepackage[pagebackref,breaklinks,colorlinks,allcolors=cvprblue]{hyperref}
\usepackage{multirow}
\def\paperID{16} 
\def\confName{CVPR}
\def\confYear{2026}

\title{GuidedFlow: An Attention-Guided Framework for Anomaly Detection in Additive Manufacturing}

\author{Sosmita Paul, Krishna Roy\\
New Mexico Institute of Mining and Technology\\
Socorro, NM 87801\\
{\tt\small 
}
}

\begin{document}
\maketitle
\begin{abstract}
Additive Manufacturing (AM) plays a vital role in the ongoing industrial revolution. However, quality control remains crucial and challenging due to printing defects or potential cyber-physical intrusions. Image or video-based anomaly detection is a key effort towards addressing these challenges. Various approaches have been explored in this domain, including reconstruction-based, embedding-based, and flow-based methods. Though normalizing flow-based methods address some of the core challenges of unforeseen defects and generalization while maintaining detection performance, existing approaches struggle with tiny/stringing defects common in 3D printing. In a small-data setting, this poses a limitation in generalization. To address these limitations, we propose \textbf{GuidedFlow}, a novel attention-guided normalizing flow model for anomaly detection and localization. GuidedFlow employs a pre-trained ResNet model, fine-tuned on the domain dataset. An attention-guided spatial and temporal flow framework models the dynamics across multiple scales and frames. A Spatio-Temporal Attention Network (SAN) enables the flow model to prioritize relevant contextual cues from input frames. We evaluate GuidedFlow on our AM3D-AD dataset, consisting of benign and anomalous real 3D printed object images and videos. We also conduct a comparative study using the MVTec-AD industrial image anomaly detection dataset. Experimental results demonstrate that GuidedFlow outperforms most of the state-of-the-art models with enhanced detection accuracy and AUROC.
\end{abstract}    
\section{Introduction}

In modern industrial environments, particularly in additive manufacturing (AM), ensuring the reliability and safety of printed components remains crucial and challenging due to printing defects or potential cyber-physical intrusions~\cite{haleem2019additive}. Even minor surface flaws or internal structural defects can cause part failure, increased production costs, or safety risks. Consequently, reliable anomaly detection and quality assurance mechanisms are critical to the safe deployment of AM technologies. In this domain, video/image-based anomaly detection offers the advantage of capturing spatio-temporal patterns of defects, making it well-suited for monitoring dynamic processes during 3D printing.

Several video/image anomaly detection and localization approaches have been explored for industrial and manufacturing applications. Reconstruction-based methods~\cite{bergmann2018improving,matsubara2020deep,venkataramanan2020attention,zavrtanik2021draem,livernoche2023diffusion,xu2023unsupervised}, such as autoencoders, generative adversarial networks (GANs) and diffusion models, learn to reconstruct benign samples and flag images with a high reconstruction error as anomalous. However, they often miss subtle or localized defects, especially when biased towards dominant regular patterns. Embedding-based methods~\cite{roth2022towards,bergmann2020uninformed} aim to learn compact latent representations, modeling normal feature distributions and detecting deviations through feature similarity. While these approaches improve generalization, they may struggle in detecting context-dependent anomalies. Recently, flow-based methods~\cite{rudolph2022fully,gudovskiy2022cflow,yu2021fastflow,lei2023pyramidflow}, particularly those based on normalizing flows, have attracted attention for their ability to estimate exact likelihoods and retain invertibility. These methods are well-suited for detecting unforeseen anomalies, as they model the full data distribution rather than just its lower-dimensional embedding.
Despite recent advancements of existing flow-based approaches, they often fall short in detecting irregular anomalies, such as stringing or tiny layer inconsistency defects that are especially common in AM processes. These problems get worse in small-data settings, where learning feature distributions is difficult. Furthermore, most existing methods treat images or frames independently and do not incorporate temporal dynamics, which is often critical for accurate localization and early detection.

To overcome these limitations, we propose GuidedFlow, a novel attention-guided normalizing flow framework for anomaly detection in additive manufacturing. GuidedFlow integrates a ResNet-50 model with a spatio-temporal attention network to extract relevant contextual features. GuidedFlow introduces a spatio-temporal attention guided flow model, enabling it to handle context-dependent anomalies. The multi-level flow blocks model appearance and structure across resolutions. By learning where and when to focus in the visual stream, GuidedFlow enhances sensitivity to small and localized defects. We evaluated GuidedFlow on our AM3D-AD dataset containing both benign and anomalous 3D-printed object images and videos. Additionally, we benchmark our model against state-of-the-art methods on the publicly available MVTec-AD ~\cite{bergmann2019mvtec} dataset, due to the limited availability of additive manufacturing datasets. The model demonstrates robust performance in both anomaly detection and localization with improved generalization compared to most of the baseline methods. 

We summarize our contributions as follows: (i) we propose and develop \textbf{GuidedFlow}, a novel conditioned-flow framework for image and video anomaly detection and localization integrating attention-guided spatial and temporal flow; (ii) we introduce a spatio-temporal attention network (SAN) using multi-head self-attention to enhance sensitivity to spatial and contextual anomalies; (iii) we propose a real-world additive manufacturing anomaly detection dataset, \textbf{AM3D-AD}, containing 3D-printed object images and videos; and (iv) we perform a comprehensive evaluation of GuidedFlow on both the AM3D-AD and MVTec-AD datasets.

\section{Related Work}

In this section, we review the key approaches for anomaly detection in AM, highlighting their advantages and shortcomings.

\subsection {Unsupervised Deep Learning Methods}

Unsupervised deep learning methods, particularly Autoencoders and Generative Adversarial Networks (GAN) based models, have gained popularity in addressing the challenge of limited anomalous data samples. In this domain, reconstruction-based strategies have shown effectiveness. For instance, MemAE, a memory augmented autoencoder, was used to detect anomalies in visual data by learning benign patterns and identifying deviations during reconstruction~\cite{gong2019memorizing}. Similarly, encoder-decoder architectures trained exclusively on defect-free data have been effective in highlighting unexpected patterns through reconstruction loss~\cite{tan2019encoder}, though they often struggle with borderline or ambiguous cases. 

Several unsupervised efforts combined traditional image features, such as oriented gradient histograms, with physics-based rendering to create open-source, layer-wise monitoring systems~\cite{petsiuk2022towards}. Such systems may be less adaptive to unseen defect variations. In~\cite{li2025application}, video data from wire arc AM systems was processed using unsupervised deep models for real-time anomaly detection, demonstrating scalability to high-throughput environments but requiring substantial computational resources.

Although existing studies emphasized the growing importance of deep unsupervised techniques with high adaptability and precision for anomaly detection in industrial settings~\cite{pang2021deep,deshpande2024deep}, generalization remains a challenge, particularly under noisy or dynamically changing manufacturing conditions.

\subsection{Transformer-Based and Contrastive Methods}

Vision Transformers (ViT) introduced a new direction for anomaly detection by modeling long-range dependencies through self-attention mechanisms~\cite{dosovitskiy2020image}. VT-ADL, a transformer-based framework, was developed to detect and locate anomalies in industrial images using the MVTec dataset~\cite{mishra2021vt}. Another work, EVAL, an explainable video anomaly localization framework, extended this by integrating explainable visual reasoning into a video anomaly localization pipeline, improving transparency in AM monitoring~\cite{singh2023eval}. To enhance robustness across varying AM processes, GeneralAD applied attention to distorted features, offering cross-domain adaptability~\cite{strater2024generalad}.

Contrastive learning approaches have also gained attention for anomaly detection. CLEP adopted contrastive pretraining for vulnerability detection, demonstrating the broader utility of discriminative representations~\cite{chen2025clep}. In the context of AM, One-for-All utilized masked cross-class contrastive learning for unsupervised anomaly detection, highlighting scalability across diverse defect types~\cite{yao2023one}. DualAD further advanced this by employing a dual-branch design to separately model normal and anomalous behavior, yielding better detection granularity~\cite{he2025dualad}. Recently, a multi-scale framework was proposed that captures both coarse and fine-grained spatio-temporal features, boosting performance in complex AM scenarios~\cite{zhang2024multi}. Despite their elevated anomaly detection performance, these methods demand substantial training data and computation, which limits real-time applicability in AM environments.

\subsection{Normalizing Flow and Diffusion Methods}
Flow-based models are increasingly being applied in anomaly detection due to their ability to model exact data likelihoods through invertible transformations. SPADE~\cite{roth2022towards}, introduced spatially-weighted normalizing flows over pretrained feature embeddings, allowing effective localization of anomalies in industrial images. Another work, PatchSVDD-Flow~\cite{yi2020patch}, embedded flow-based priors into support vector data descriptions, resulting in more compact latent distributions for benign data. FastFlow~\cite{yu2021fastflow} used 2D normalization flows for efficient unsupervised detection and localization of anomalies in visual data. In the temporal domain, Lu et al.~\cite{liu2018future} proposed a future frame prediction model that employs normalizing flows over temporal embeddings, demonstrating its applicability in video-based surveillance anomaly detection. Recently, flow-based models have been extended toward spatio-temporal domains using hierarchical and multi-scale architectures. PyramidFlow Matching~\cite{jin2024pyramidal} implements a coarse-to-fine flow matching strategy across Laplacian pyramid levels to model fine-grained spatial and motion cues. This is a diffusion-based approach utilizing a unified Diffusion Transformer (DiT) for video generation modeling. Diffusion-based methods\cite{livernoche2023diffusion,xu2023unsupervised,jin2024pyramidal} offer strong generative power for reconstructing normal data in anomaly detection. Although diffusion-based models are effective in capturing complex structures, they are computationally intensive and may blur fine details critical to accurate stringing defect localization.

Flow-based models may still encounter limitations in detecting tiny printing defects, such as stringing in 3D printing, causing a generalization challenge. Likelihood-based scoring can be misled by high-frequency background patterns, and pretrained features may lack semantic alignment with task-specific distributions. To address these challenges, we propose \textbf{GuidedFlow}, an attention-guided, hierarchical flow framework that integrates spatio-temporal guidance into the flow levels.

\begin{figure*}[!htpb]
  \centering
  \begin{minipage}[b]{0.9\textwidth}
    \includegraphics[width=\textwidth]{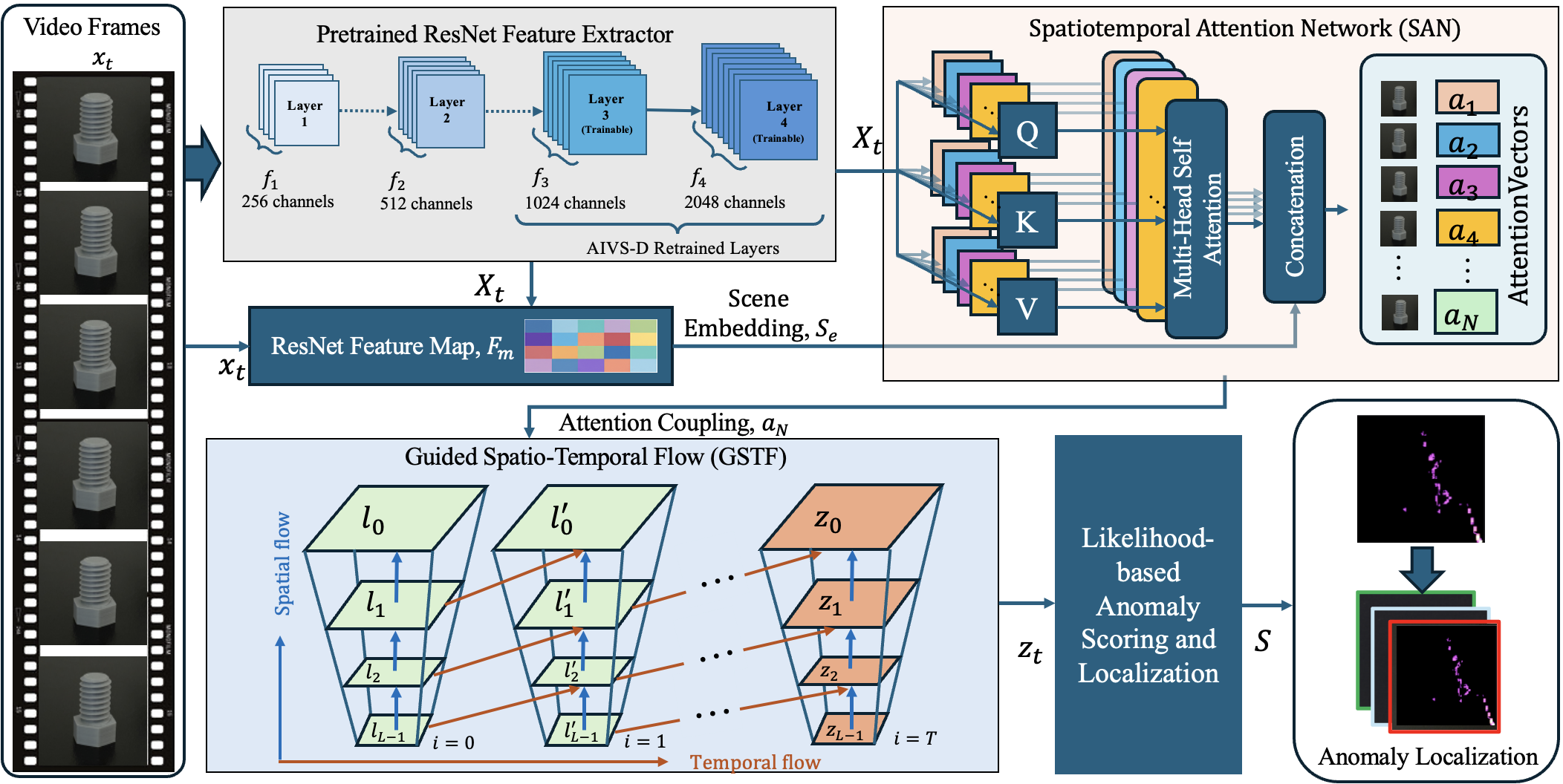}
    \caption{GuidedFlow Overview. The model is composed of four parts: ResNet Feature Extractor, Spatio-temporal Attention Network, Guided Spatiot-Temporal Flow, and Anomaly Localization Map.}
    \label{fig:system_architecture}
  \end{minipage}
\end{figure*}

\section{Methodology}

The proposed \textbf{\textit{GuidedFlow}} system architecture is shown in Figure~\ref{fig:system_architecture}. We introduce the GuidedFlow model, a deep framework for high-resolution video anomaly detection and localization using an attention guided spatio-temporal flow mechanism. It models the distribution of multi-scale feature pyramids extracted from frames rather than raw pixels. The model is trained on benign data using a likelihood objective. GuidedFlow processes $224 \times 224$, 3-channel video clips and identifies 3D printing defects, followed by defect localization. The approach uses a pretrained feature extractor, spatio-temporal attention mechanisms, attention-guided multi-scale flow modeling for precise localization and detection of spatio-temporal anomalies.



\subsection{ResNet Feature Extraction}
A pretrained ResNet is used to extract video features. Given a video frame $x_t \in \mathbb{R}^{3 \times H \times W}$, each frame is first passed through a ResNet-50 backbone to extract hierarchical features $X_t = E(x_t)$ from the first 4 ResNet layers named $f_1, f_2, f_3, f_4$ with channel depths of 256, 512, 1024, and 2048 respectively. During training, layers $f_1$, $f_2$ were frozen to preserve general low-level feature representations and layers $f_3$, $f_4$ are trained. 
The extracted features $X_t$, along with the original frame $x_t$ and feature map $F_m$ are combined to generate a scene embedding $S_e$.
The ResNet feature map $F_m$ captures spatial information across varying resolutions. Extracted features $X_t$ and the scene embeddings $S_e$ are fed to the spatio-temporal attention network to the next stage.

\subsection{SAN Spatio-temporal Attention Encoding}
To capture spatial and temporal dependencies across video frames and prioritize potential mapping of the anomaly, a spatio-temporal attention module is applied on the ResNet features. We used a multi-head self-attention mechanism, where the input $X_t$ is transformed into query $Q$, key $K$, and value $V$ vectors via linear projections, with the output computed as: $\alpha_N =\text{Attention}(Q, K, V)$. The attention mechanism is defined as:

\begin{equation}
    \text{Attention}(Q, K, V) = \text{softmax}\left(\frac{QK^T}{\sqrt{d_k}}\right)V,
\end{equation}
Where $( d_k )$ denotes the dimension of the key vector. The network implements multiple attention heads (H heads), each processing a subset of $X_t$ independently. The outputs of these heads $\alpha_N$ and scene embeddings $S_e$ from the previous stage are concatenated and linearly transformed to produce a unified attention output vector $a_1$ to  $a_N$ for N frames:

\begin{equation}
    a_N = \text{cat}(S_e, \text{head}_1, \ldots, \text{head}_H) W_o
\end{equation}
where $ W_o $ is an output projection matrix, and $ H $ is the number of heads. $a_N$ reflects the spatial and temporal dependencies and enhances the ability to focus on relevant regions across frame sequences in the GuidedFlow mechanism. Thus, the SAN module generates the spatio-temporal context embedding for each frame, $a = C_{\phi}(X_t, S_e)$, where $C_{\phi}(\cdot)$ is learnable and produces conditioning attention $a$.

\subsection{Guided Spatio-Temporal Flow Processing}
The ResNet features $X_t$ are passed to a Guided Spatio-Temporal Flow (GSTF) framework. The GSTF module integrates spatial and temporal flow components conditioned to SAN attention to process video features \( X_t \) through a hierarchical, invertible transformation. GSTF leverages a Laplacian pyramid decomposition and attention coupling to model spatial dependencies within frames and temporal dependencies across frames. First, we build a multi-scale Laplacian pyramid  
$\mathcal{P}(X_t) = \{lp_0, lp_1, \dots, lp_L\}$ from the backbone features and apply a conditional invertible transformation to obtain latent variables:

\begin{equation}
\mathbf{z} = F_{\theta}(\textbf{l}; a).
\label{eq:condflow}
\end{equation}

The mapping $F_{\theta}(\cdot; a)$ is invertible with respect to $\textbf{l}$ for any fixed $a$, enabling exact likelihood evaluation. The spatial and temporal flow components are explained in the next sections.

\vspace{0.5em}
\noindent \textbf{Spatial Flow:} The spatial flow is implemented to transform feature representations across multiple scales and spatial dynamics. The flow mechanism starts with the construction of a \texttt{Laplacian pyramid} from the features $X_t$. The \texttt{Laplacian Pyramid} decomposes the input features $X_t$ into \( L \) levels, where each level \( l_i \) represents a scale-specific representation, 


\begin{equation}
l_i' = f_{\text{spatial}}(l_i, a),
\end{equation}
where \( f_{\text{spatial}} \) are coupling across frames. The spatial flow applies invertible transformations to each of the pyramid levels. The spatial flow thus transforms the pyramid levels \( l_i \) to \( l_i' \) across scales, guided by context coupling \( a \), with invertibility ensured by the coupling and convolution operations.

\noindent \textbf{Temporal Flow:} The temporal flow extends the spatial transformations across the temporal dimension \( T \), modeling dependencies between frames \( i = 0 \) to \( i = T \). The input to the temporal flow is the spatially transformed pyramid levels $ l_i'  \in \mathbb{R}^{B \times C \times T \times H_l \times W_l} $ for each level $l$. The temporal flow is modeled as a sequence of invertible coupling operations across $ l_i'$, spatio-temporal attention vectors $a$:
\begin{equation}
    l_i'' = f_{\text{temporal}}(l_i', a),
\end{equation}

Where \( f_{\text{temporal}} \) represents the temporal flow function, implemented through the invertible coupling.

Each GSTF layer uses affine coupling. Splitting an input $u=[u_{1},u_{2}]$, we define
\begin{equation}
v_{1} = u_{1}, \qquad
v_{2} = u_{2}\odot \exp(s(u_{1};a)) + t(u_{1};a),
\label{eq:coupling}
\end{equation}
where $(s(\cdot;a), t(\cdot;a))$ are produced by a conditioning network.
We implement conditioning via a hypernetwork modulation:
\begin{equation}
[s,t] = g(u_{1}) \;\oplus\; h(a),
\label{eq:film}
\end{equation}
where $g(\cdot)$ is the standard coupling subnetwork and $h(\cdot)$ maps SAN context $a$ to additive (or scale/shift) modulation of coupling parameters.
The log-determinant is computed as
\begin{equation}
\log\left|\det\frac{\partial v}{\partial u}\right| = \sum \, s(u_{1};a).
\label{eq:logdet}
\end{equation}

The spatial flow transforms pyramid levels using scale-specific invertible coupling and convolution, while the temporal flow extends this across frames using 3D coupling guided by spatio-temporal attention $a$. The invertibility of the GSTF flow mechanism makes GuidedFlow computationally efficient for fast likelihood based anomaly scoring and localization.




\subsection{Anomaly Scoring and Localization}
For anomaly detection, we calculate anomaly score $S(x_t)$ from $\mathbf{l}=\mathcal{P}(E(x_t))$ and $a=C_{\phi}(E(x_t))$ by evaluating the negative log-likelihood:
\begin{equation}
S(x) = -\log p(\mathbf{l}\mid a).
\label{eq:score}
\end{equation}

We compute per-location contributions from the latent energy and log-determinant terms, aggregate across pyramid levels, and upsample to the input resolution:
\begin{equation}
M(x_t) = \sum_{j=0}^{L-1}\uparrow_{H,W}\Big(
\sum_{c}\tfrac{1}{2}(z^{j}_{cuv})^{2} - d^{j}_{uv}
\Big),
\label{eq:heatmap}
\end{equation}
where $d^{j}_{uv}$ denotes the spatially-resolved log-determinant contribution induced by the coupling layers at level $j$.


\section{Experiment}
 \subsection{Experimental Settings}
\subsubsection{Dataset Details} To evaluate the effectiveness of the GuidedFlow framework, we conduct experiments on a domain dataset AM3D-AD (ours) and a subset of the baseline MVTec-AD ~\cite{bergmann2019mvtec} dataset. 

\begin{figure}[!h]
    \centering
    \begin{subfigure}{0.473\linewidth}
        \centering
        \includegraphics[width=\linewidth]{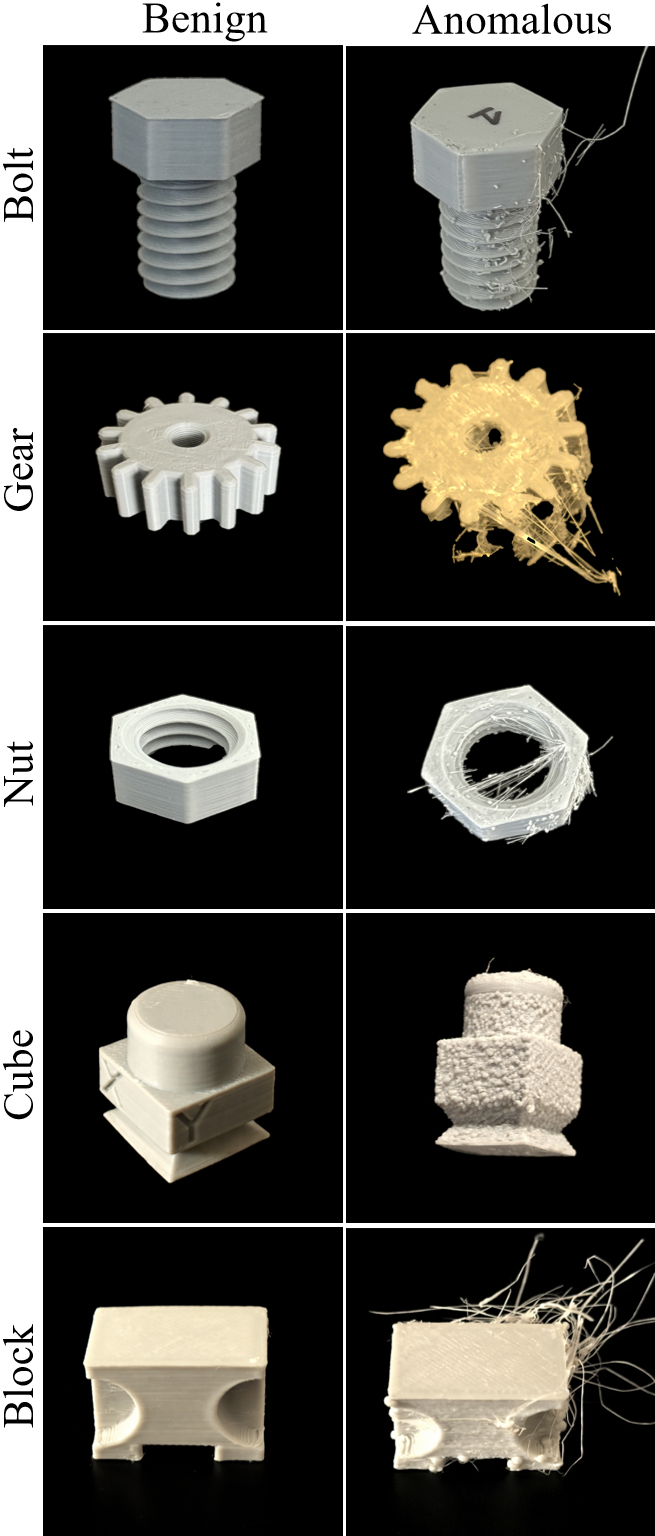}
        \caption{AM3D-AD Dataset}
        \label{fig:am3d}
    \end{subfigure}
    \hfill
    \begin{subfigure}[b]{0.487\linewidth}
        \centering
        \includegraphics[width=\linewidth]{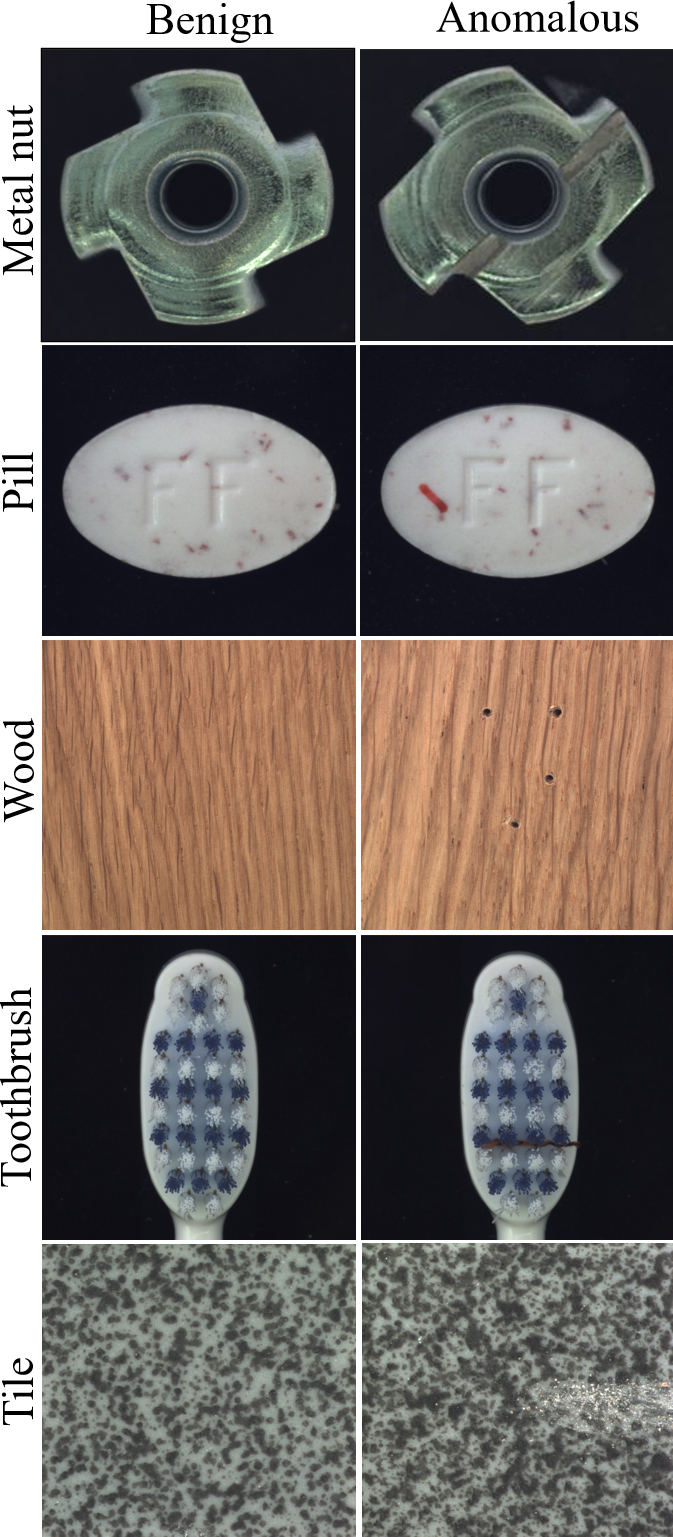}
        \caption{MVTec-AD Dataset}
        \label{fig:mvtec}
    \end{subfigure}
    \caption{AM3D-AD and MVTec-AD dataset samples. The first column shows benign samples, while the second column shows anomalous samples in each dataset.}
    \label{fig:dataset_fig}
\end{figure}

\textbf{AM3D-AD Dataset} contains both images and videos from five categories: Bolt, Gear, Nut, Block, and Cube. All the objects are printed with NylonX Carbon Fiber. The dataset consists of $537$ images ($254$ benign and $283$ anomalous) and $129$ videos of the printed object, each $25$ seconds long on average, supporting temporal analysis. To create anomalous samples with 3D printing defects, the G-code was intentionally sabotaged by inserting redundant or noisy motion commands, including high-frequency or random G0 (non-extruding move) and G1 (extruding move) instructions. These modifications cause surface-level distortions by introducing non-uniform paths or jagged trajectories. The attacks were performed in a controlled environment with Creality K1 Max and Ender-3 3D printers\cite{creality}. Further details regarding the anomalous data collection procedure can be found in \cite{hasan2026engineering}. All the images and videos are manually labeled. An overview of the AM3D dataset is shown in Figure~\ref{fig:am3d}.



\textbf{MVTec-AD Dataset} comprises 5,354 high-resolution images across 15 distinct categories, with a training set of 3,629 anomaly-free images and a test set of 1,725 images containing normal and anomalous samples. 
An illustration is shown in Figure~\ref{fig:mvtec}.


\textbf{Evaluation Metrics} We evaluate our method using the Area Under the Receiver Operating Characteristic Curve (AUROC) at the pixel P-AUROC and image level I-AUROC for image anomaly detection and localization. Video-level AUROC is reported as V-AUROC. We also report the Area Under the Per-Region Overlap (AUPRO) curve and the Average Precision (AP). AUROC scores are reported in percentage (\%). 



\textbf{Implementation Details} 
GuidedFlow is implemented using Pytorch 2.5 in Python 3.10. Our implementation uses ResNet-50. The spatio-temporal attention network was implemented with a multi-head self-attention encoder with $H = 8$ attention heads, chosen according to the hyperparameter performance analysis discussed later. The guided-flow mechanism is implemented with a spatial flow and a temporal flow containing $6$ flow levels $4$ temporal flow stacks.


\subsubsection{Training and Evaluation Setup}
Model training and comprehensive evaluation are performed in multiple parameter settings, including an ablation study. 

\textbf{Training Details:} In the data pre-processing phase, all images from AM3D-AD and MVTec-AD datasets are resized to 224x224. In training, we assume a standard normal prior to latent variables $\mathbf{z}$:
\begin{equation}
p(\mathbf{z}) = \prod_{j=0}^{L-1}\mathcal{N}(z^{j};0,I).
\end{equation}
The conditional log-likelihood of the pyramid representation is
\begin{equation}
\log p(\mathbf{l}\mid a)=\sum_{j=0}^{L-1}\left(\log p(z^{j})+\log\left|\det\frac{\partial F^{j}_{\theta}}{\partial l^{j}}\right|\right).
\label{eq:condlik}
\end{equation}

We train the flow model using benign samples only by minimizing the negative log-likelihood:
\begin{equation}
\mathcal{L}_{\text{NLL}}
= -\mathbb{E}_{x_t\sim\mathcal{D}_{\text{benign}}}\Big[
\log p\big(\mathcal{P}(E(x_t)) \mid C_{\phi}(E(x_t))\big)
\Big].
\label{eq:nll_final}
\end{equation}
During training, the backbone $E(\cdot)$ is trained, while SAN parameters $\phi$ and flow parameters $\theta$ are optimized jointly.

The complete framework is trained on benign samples of both datasets in two phases: Phase I) Backbone training: Top layers of the ResNet or $1 \times 1$ Convolution feature extractor trained with an Adam optimizer, maintaining a learning rate of $1\times10^{-4}$, batch size of 8, and training for 20 epochs. Phase-II) Conditioned Spatio-Temporal Flow Learning: The SAN and GSTF flow model is trained with an invertible normalizing flow model, maintaining a learning rate of $1\times10^{-4}$, a batch size of 8, and training for 100 epochs.

\begin{table*}[!h]
\centering
\caption{Comparison of image anomaly detection methods on AM3D and MVTec datasets using I-AUROC, P-AUROC, AUPRO, and I-AP.}
\footnotesize
\begin{tabular}{l|l|ccccc|ccccc|c}
\hline
\multirow{2}{*}{\textbf{Method}} & \multirow{2}{*}{\textbf{Metric}} & \multicolumn{5}{c|}{\textbf{AM3D Dataset}} & \multicolumn{5}{c|}{\textbf{MVTec Dataset}} & \multirow{2}{*}{\textbf{Avg}} \\
 & & Bolt & Nut & Gear & Cube & Block & MetalNut & Pill & Toothbrush & Tile & Wood & \\
\hline
\multirow{3}{*}{DiffusionAD} 
  & I-AUROC   & 97.8 & 97.3 & 96.9 & 94.9 & 95.9  & 99.1 & 98.3 & 95.7 & 96.2 & 96.0 & 96.8 \\
  & P-AUROC   & 98.1 & 96.9 & 95.5 & 93.5 & 92.6  & 97.8 & 97.3 & 94.1 & 96.1 & 95.9 & 95.8 \\
  & AUPRO & 0.92 & 0.86 & 0.84 & 0.79 & 0.85  & 0.87 & 0.86 & 0.82 & 0.91 & 0.88 & 0.86 \\
   & I-AP  & 0.92 & 0.87 & 0.91 & 0.84 & 0.87 & 0.90 & 0.85 & 0.82 & 0.84 & 0.86& 0.87\\
\hline
\multirow{3}{*}{INP-Former} 
  & I-AUROC   & 96.5 & 96.4 & 97.2 & 94.0 & 94.6  & 97.5 & 96.3 & 94.1 & 94.0 & 95.4 & 95.6 \\
  & P-AUROC   & 97.2 & 96.6 & 96.5 & 93.8 & 93.5  & 96.8 & 96.1 & 93.8 & 93.5 & 93.8 & 95.2 \\
  & AUPRO & 0.93 & 0.92 & 0.86 & 0.83 & 0.91  & 0.92 & 0.94 & 0.85 & 0.92 & 0.91 & 0.90 \\
   & I-AP  & 0.95 & 0.94 & 0.91 & 0.89 & 0.92 & 0.91 & 0.94 & 0.89 & 0.88 & 0.92 & 0.92 \\
\hline
\multirow{3}{*}{PyramidFlow} 
  & I-AUROC   & 97.5 & 96.4 & 95.7 & 94.9 & 95.8 & 98.1 & 97.4 & 93.9 & 95.7 & 95.3 & 96.1 \\
  & P-AUROC   & 96.2 & 95.6 & 94.7 & 93.8 & 94.0 & 97.0 & 95.8 & 94.9 & 95.2 & 95.3 & 95.3 \\
  & AUPRO & 0.94 & 0.94 & 0.89 & 0.86 & 0.92  & 0.91 & 0.92 & 0.87 & 0.91 & 0.95 & 0.91 \\
   & I-AP  & 0.97 & 0.95 & 0.94 & 0.92 & 0.93 & 0.95 & 0.96 & 0.89 & 0.94 & 0.93 & 0.94 \\
\hline
\multirow{3}{*}{GuidedFlow} 
  & I-AUROC   & 98.7 & 98.2 & 97.9 & 95.5 & 96.0  & 99.3 & 98.9 & 95.5 & 97.1 & 96.7 & 97.4 \\
  & P-AUROC   & 98.9 & 97.8 & 96.7 & 94.2 & 94.1  & 98.8 & 98.1 & 95.8 & 96.4 & 96.8 & 96.8 \\
  & AUPRO & 0.96 & 0.93 & 0.91 & 0.88 & 0.94  & 0.93 & 0.94 & 0.89 & 0.96 & 0.96 & 0.93 \\
   & I-AP  & 0.97 & 0.97 & 0.96 & 0.94 & 0.95 & 0.98 & 0.98 & 0.93 & 0.93 & 0.95 & 0.95 \\
\hline
\end{tabular}
\label{tab:method_comparison_avg}
\end{table*}

\begin{table}[!h]
\centering
\caption{GuidedFlow video anomaly detection performance on AM3D dataset via video-level AUROC (V-AUROC), P-AUROC, AUPRO, and AP.}
\footnotesize
\begin{tabular}{l|ccccc}
\hline
\multirow{2}{*}{\textbf{Metric}} & \multicolumn{5}{c}{\textbf{AM3D Dataset}} \\
  & Bolt & Nut & Gear & Cube & Block \\
\hline
  V-AUROC   & 98.9 & 98.8 & 98.9 & 97.2 & 98.6  \\
  P-AUROC   & 99.1 & 98.7 & 98.2 & 96.4 & 96.1  \\
  AUPRO  & 0.94 & 0.93 & 0.93 & 0.92 & 0.95  \\
  AP  & 0.97 & 0.96 & 0.97 & 0.94 & 0.94  \\
\hline
\end{tabular}
\label{tab:vid_detection}
\end{table}


\paragraph{Evaluation Details}
For both datasets, we split the benign set into 80\% for training and 20\% for testing. All anomalous samples are used only at test time.
All model training and evaluation are performed on a NVIDIA A100 GPU server with 40 GB of memory.

\subsection{Anomaly Detection Performance}
We calculate AUROC at the image and pixel levels presented with I-AUROC and P-AUROC, respectively, to evaluate the performance of GuidedFlow without temporal-flow (T-Flow) and temporal-attention (T-Attn) anomaly detection. It achieves I-AUROC and P-AUROC scores, ranging from 95.5 to 99.3 and 94.1 to 98.9, respectively, with AUPRO between 0.88 and 0.96 and I-AP ranging from 0.93 to 0.98 on the two datasets. GuidedFlow I-AUROC outperforms for the majority classes with Bolt in the AM3D dataset and MetalNut in the MVTec dataset, achieving the highest scores of 98.7 and 99.3, respectively. GuidedFlow effectively generalizes across the 10 classes, with consistent AUROC scores. The results are presented in Table~\ref{tab:method_comparison_avg}.

\begin{figure}[!h]
    \centering
    \begin{subfigure}{\linewidth}
        \centering
        \includegraphics[width=\linewidth]{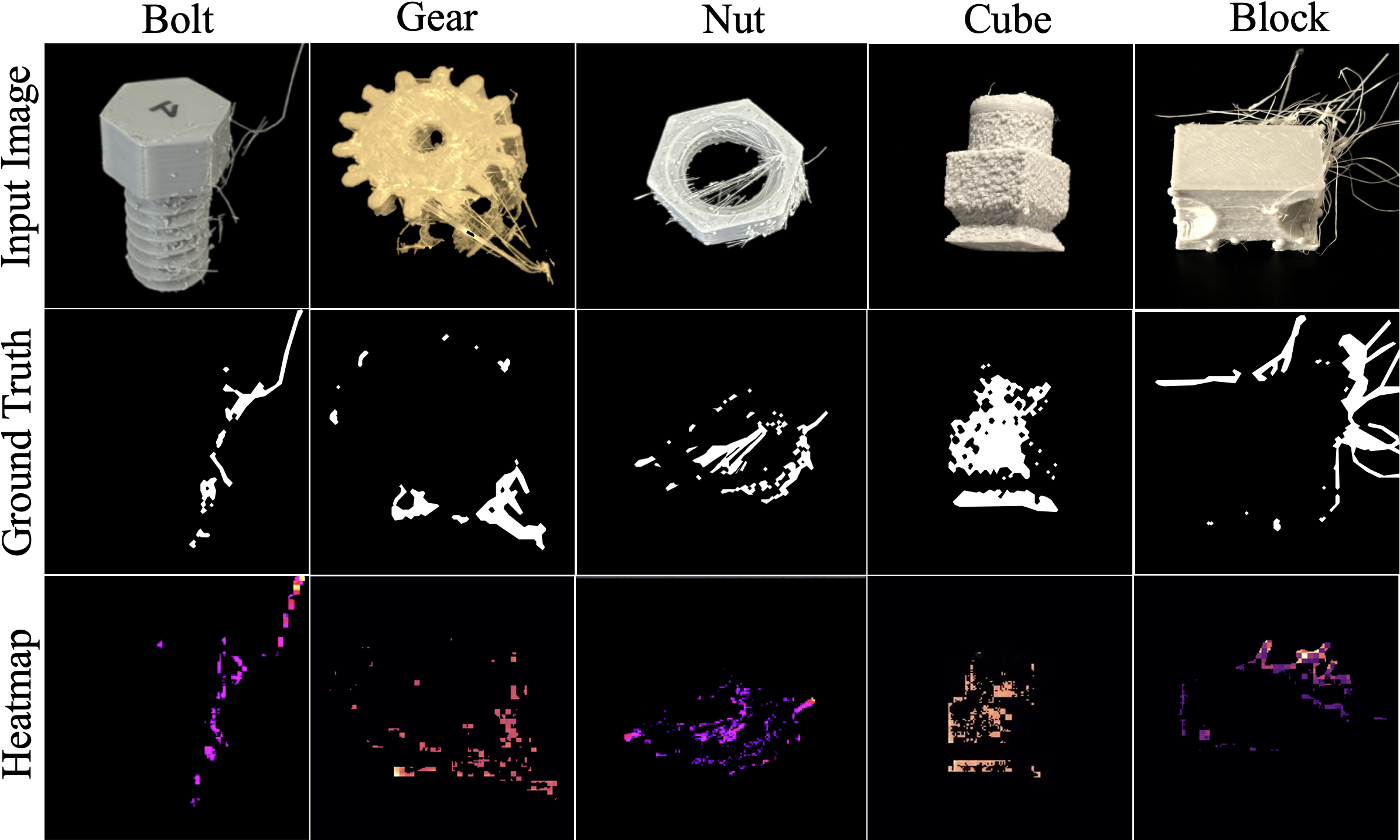}
        \caption{AM3D-AD Anomaly Visualization}
        \label{fig:am3dheat}
    \end{subfigure}
    \hfill
    \begin{subfigure}[b]{\linewidth}
        \centering
        \includegraphics[width=\linewidth]{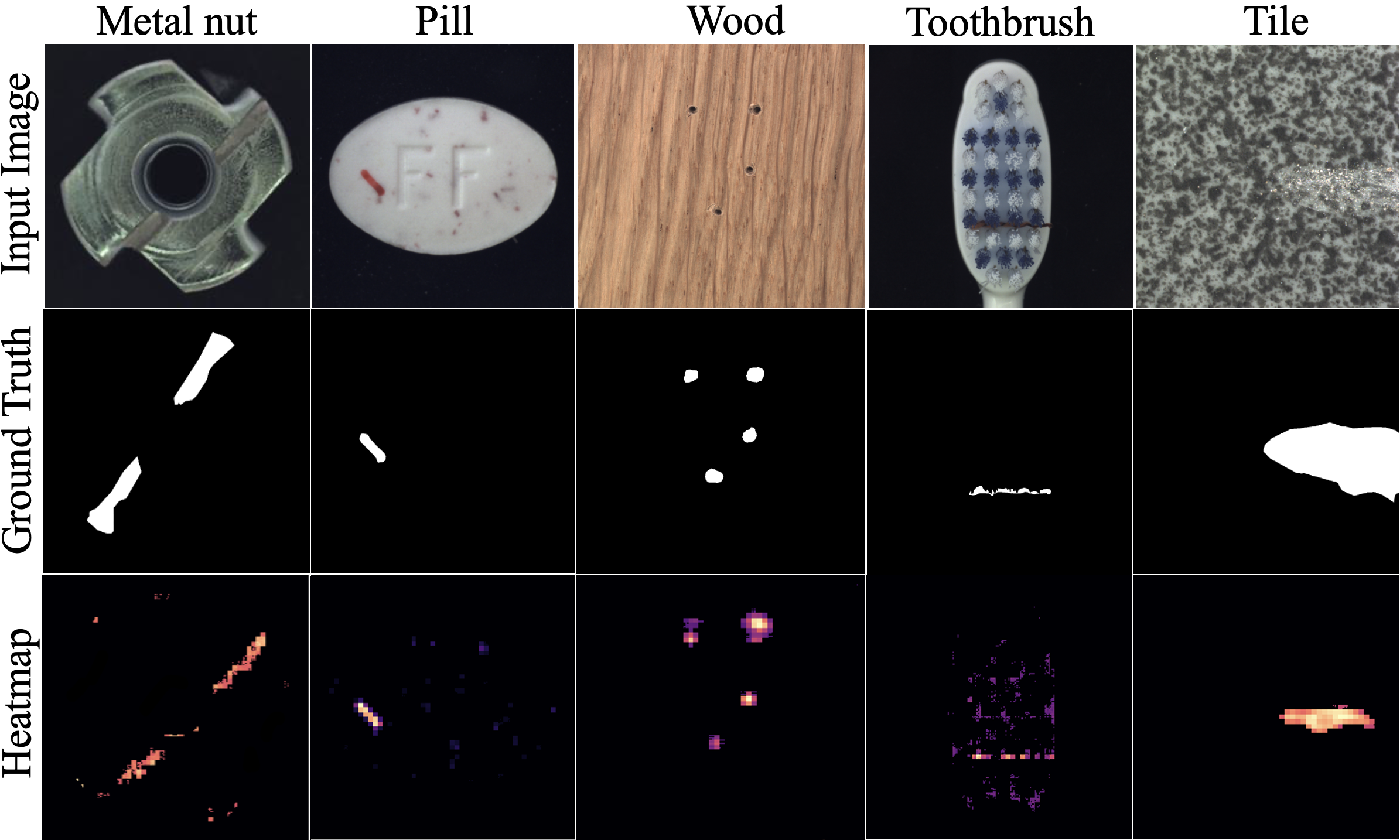}
        \caption{MVTec-AD Anomaly Visualization}
        \label{fig:mvtecheat}
    \end{subfigure}
    \caption{Visualization of GuidedFlow anomaly localization results on AM3D-AD and MVTec-AD datasets. The top row is input images, the middle row ground truths, and the bottom row GuidedFlow localization as heatmaps, where higher intensity values indicate a higher anomaly likelihood}
    \label{fig:heatmap_fig}
\end{figure}

We performed a comparative study of GuidedFlow against three state-of-the-art (SOTA) anomaly detection models, including DiffusionAD~\cite{zhang2025diffusionad}, INP-Former~\cite{luo2025exploring} and PyramidFlow~\cite{lei2023pyramidflow} as shown in Table~\ref{tab:method_comparison_avg}. All baseline models are trained with the respective AM3D-AD and MVTec-AD datasets for fair comparison. GuidedFlow attains an average I-AUROC of 97.4, an average P-AUROC of 96.8, AUPRO 0.93, and I-AP 0.95, outperforming most of the SOTA models  DiffusionAD(96.8, 95.8, 0.86, 0.87), INP-Former (95.6, 95.2, 0.90, 0.92), and PyramidFlow (96.1, 95.3, 0.91, 0.94). The inclusion of guided attention and spatial flow enabled the model to distinguish manufacturing defects. GuidedFlow showed consistent performance across multiple categories of defects, especially in visually similar cases where baseline models struggled. However, GuidedFlow shows a lower P-AUROC for some specific classes, such as Cube (94.2) and Block (94.1) in the AM3D-AD dataset, and Toothbrush (95.8) and Tile (96.4) in the MVTec-AD dataset, potentially posing questions about the generalizability of the model. This indicates that GuidedFlow may struggle with anomalies that exhibit minimal pixel-wise deviation or complex textures like in the case of Cube (AM3D) and Toothbrush (MVTec), causing model to emphasize global texture over subtle structural defects. A visualization of the heatmaps of the AM3D-AD and MVTec-AD sample is presented in Figure~\ref{fig:heatmap_fig}.


To evaluate video anomaly detection and localization performance, we experiment with GuidedFlow, including temporal attention (T-Attention) and temporal flow (T-Flow) on the AM3D-AD video dataset. The video-level detection results are reported in Table~\ref{tab:vid_detection}. We observe that GuidedFlow significantly improves V-AUROC compared to image-level I-AUROC for several categories. In particular, the model achieves 98.9, 97.2, and 98.6 V-AUROC for Gear, Cube, and Block, respectively. These improvements can be attributed to the temporal attention mechanism of SAN that leverages contextual information from four sequential frames (see the hyperparameter analysis in Section~\ref{subsec:parameter}). GuidedFlow also achieves improved AUPRO scores (0.93 for Gear and 0.92 for Cube) compared to image-based AUPRO, while maintaining stable AP across all categories. These results demonstrate that GuidedFlow effectively captures temporal consistency and accurately detects and localizes anomalies across diverse object geometries and defect patterns, indicating strong generalization across AM3D categories.

\begin{table}[!ht]
\centering
\caption{Ablation of GuidedFlow architectural components for AM3D-AD video dataset. "S-Flow", "T-Flow" refer to the spatial and temporal flow respectively and "ST-Flow"  refer to the spatio-temporal flow components. Similarly "S-Attention or S-Attn", "T-Attention or T-Attn" refer to the spatial and temporal attention respectively, and "ST-Attention or ST-Attn" refer to the spatio-temporal attention.}
\footnotesize
\begin{tabular}{lccc}
\hline
\textbf{Model Variants} & \textbf{V-AUROC} & \textbf{P-AUROC} \\
\hline
1$\times$1Conv + ST-Flow                          & 91.5 & 89.5  \\
TunedResNet + ST-Flow                             & 92.1 & 91.0  \\
ResNet + S-Attn + ST-Flow   & 97.4 & 95.2  \\
ResNet + T-Attn + ST-Flow   & 96.1 & 94.8  \\
ResNet + ST-Attn + S-Flow   & 95.3 & 96.1  \\
ResNet + ST-Attn + T-Flow   & 94.6 & 94.8  \\
\textbf{GuidedFlow (ST-Attn + ST-Flow)} & \textbf{98.7} & \textbf{98.9}  \\
\hline
\end{tabular}
\label{tab:guidedflow_ablation}
\end{table}

\subsubsection{Qualitative Evaluation}
Fig.~\ref{fig:heatmap_fig} visualizes qualitative localization results, showing the input frame, ground-truth, and the predicted anomaly heatmap produced by GuidedFlow. In the heatmap, higher intensity indicates a higher likelihood of anomalies.

\subsection{Ablation Study}
To evaluate the significance of different architectural components of GuidedFlow, we performed an ablation study on the AM3D-AD video dataset presented in Table~\ref{tab:guidedflow_ablation}. We started with a simple 1×1 convolution connected to flow blocks without attention module SAN and observed a modest performance with an V-AUROC of 91.5 and a P-AUROC of 89.5. Replacing the shallow encoder with AM3D data-tuned ResNet-50 improves both video and pixel-level performance, suggesting that stronger semantic features help the flow model better capture benign patterns. 

\textbf{Effectiveness of Spatial and Temporal Attention} We introduce spatial attention on top of the ResNet feature extraction (S-Attention + ST-Flow) that demonstrates a significant boost in both V-AUROC and P-AUROC to 97.4 and 95.2 respectively. Spatial guidance clearly adds to the learning of contextual representations for anomaly localization.
Temporal attention (T-Attention + ST-Flow) also improves performance over the models without attention coupling (\textit{1×1Conv+ST-Flow} or \textit{TunedResNet+ST-Flow}). Moreover, combining both spatial and temporal attention in the GuidedFlow architecture leads to the highest performance, with a V-AUROC of 98.7 and P-AUROC of 98.9. These results indicate that spatial and temporal cues are complementary to each other. Spatial attention enhances feature focus across spatial hierarchies within a frame, while temporal attention reinforces consistency in sequential frame context. 


\textbf{Effectiveness of Spatial and Temporal Flow}
To assess the significance of spatial and temporal flow, we designed two variants of GuidedFlow, \textit{ST-Attention + S-Flow} and \textit{ST-Attention + T-Flow}, varying the type of normalizing flow being applied and trained with the AM3D-AD video dataset. The ablation results for these two variants in Table~\ref{tab:guidedflow_ablation} shows complementary strengths of spatial and temporal flow mechanisms. Both \textit{ST-Attention + S-Flow} and \textit{ST-Attention + T-Flow} achieve higher V-AUROC of 95.3 and 94.6, indicating the effectiveness of spatial and temporal flow in capturing spatial and sequence-level deviations during video anomaly detection. Multi-resolution spatial and multi-frame temporal encoding improve the model’s sensitivity to localized fine/stringing defects. While temporal flow improves overall detection performance, spatial flow contributes more directly to accurate and fine-grained anomaly localization. In GuidedFlow the spatial flow obtains strong detection performance, and the temporal flow further improves it by incorporating inter-frame relational context.

\subsection{Parameter Performance}
\label{subsec:parameter}
\begin{figure}[!hptb]
    \centering
    \includegraphics[width=\linewidth]{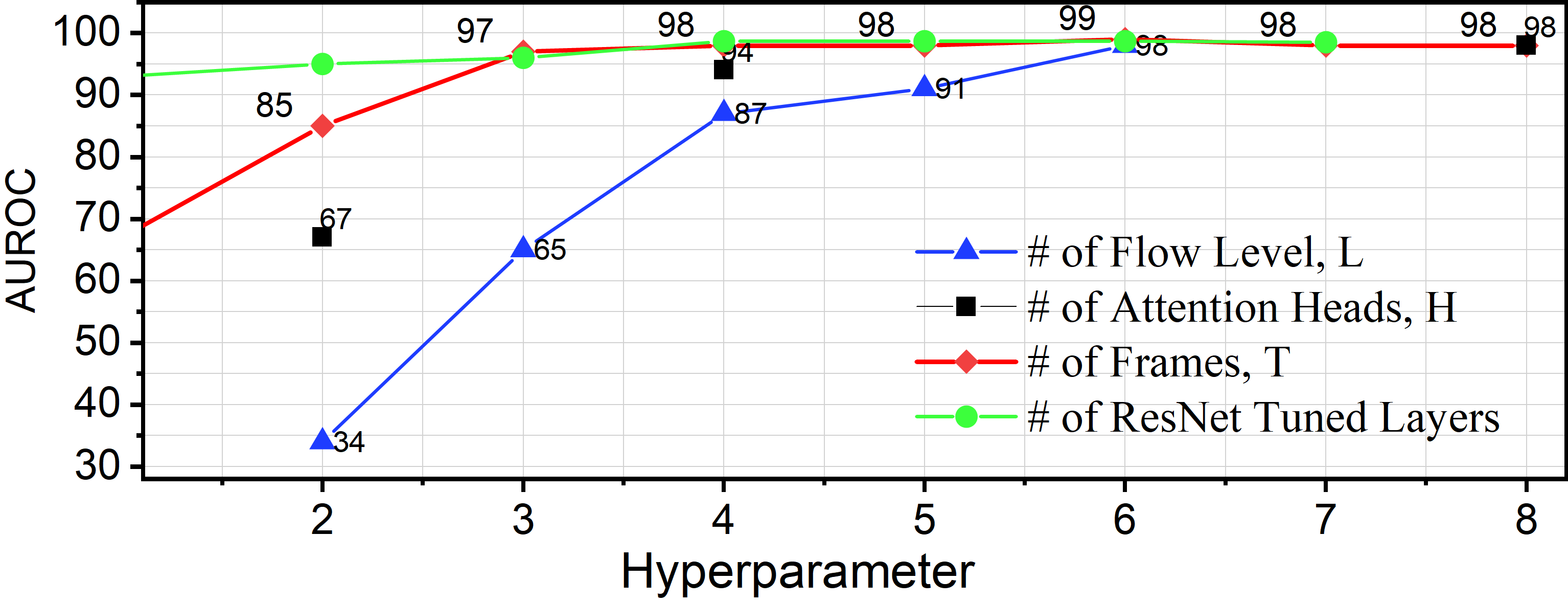}
    \caption{GuidedFlow hyperparameter performance. AUROC values for the hyperparameter settings are shown as percentages (\%). }
    \label{fig:hyperpara}
\end{figure}

We evaluate GuidedFlow’s performance using different hyperparameter settings with the number of flow levels $L$, number of temporal frames $T$, number of attention heads $H$, and number of ResNet Tuned Layers. A detailed hyperparameter vs. AUROC analysis is shown in Figure~\ref{fig:hyperpara}. We calculate AUROC for different combinations of $L$, $T$, $H$, and Tuned Layers.

We observe that the number of flow levels $L = 6$, the temporal frames $T = 3$, and the attention heads $H = 8$ produce the highest AUROC with four tuned ResNet layers.

\section{Limitations and Future Work}
While GuidedFlow demonstrates strong overall performance, we observe reduced effectiveness for certain classes, such as Block and Cube in AM3D-AD, and Toothbrush and Tile in MVTec-AD, where anomalies show minimal pixel-wise deviation or highly repetitive textures. In addition, evaluation is currently limited to a small number of publicly available additive manufacturing datasets. Future work will expand experiments to more domain datasets, including diverse materials and printing conditions, and explore improved domain-adaptive feature learning to enhance robustness and generalizability.


\section{Conclusion}
In this work, we introduced GuidedFlow, an attention-guided spatio-temporal normalizing flow model designed for detecting and localizing anomalies in both images and videos in additive manufacturing. The model uses hierarchical features from a pretrained ResNet-50 with spatio-temporal attention conditioned flow processing, enabling analysis across multiple resolutions and frames. We evaluated the model on our AM3D-AD dataset, which includes real images and videos of 3D-printed objects, and the MVTec-AD baseline anomaly detection dataset. Experimental results demonstrate consistent performance in both image-based and video-based anomaly detection tasks.

\section*{Acknowledgements}
This material is based upon work supported by the National Science Foundation under Award No. 2417062. Any opinions, findings, and conclusions or recommendations expressed in this material are those of the authors and do not necessarily reflect the views of the National Science Foundation.



{
    \small
    \bibliographystyle{ieeenat_fullname}
    \bibliography{cvpr_vision}

@inproceedings{lei2023pyramidflow,
  title={Pyramidflow: High-resolution defect contrastive localization using pyramid normalizing flow},
  author={Lei, Jiarui and Hu, Xiaobo and Wang, Yue and Liu, Dong},
  booktitle={Proceedings of the IEEE/CVF conference on computer vision and pattern recognition},
  pages={14143--14152},
  year={2023}
}

@inproceedings{gong2019memorizing,
  title={Memorizing normality to detect anomaly: Memory-augmented deep autoencoder for unsupervised anomaly detection},
  author={Gong, Dong and Liu, Lingqiao and Le, Vuong and Saha, Budhaditya and Mansour, Moussa Reda and Venkatesh, Svetha and Hengel, Anton van den},
  booktitle={Proceedings of the IEEE/CVF international conference on computer vision},
  pages={1705--1714},
  year={2019}
}

@article{pang2021deep,
  title={Deep learning for anomaly detection: A review},
  author={Pang, Guansong and Shen, Chunhua and Cao, Longbing and Hengel, Anton Van Den},
  journal={ACM computing surveys (CSUR)},
  volume={54},
  number={2},
  pages={1--38},
  year={2021},
  publisher={ACM New York, NY, USA}
}

@article{petsiuk2022towards,
  title={Towards smart monitored AM: Open source in-situ layer-wise 3D printing image anomaly detection using histograms of oriented gradients and a physics-based rendering engine},
  author={Petsiuk, Aliaksei and Pearce, Joshua M},
  journal={Additive Manufacturing},
  volume={52},
  pages={102690},
  year={2022},
  publisher={Elsevier}
}

@article{li2025application,
  title={Application of unsupervised learning methods based on video data for real-time anomaly detection in wire arc additive manufacturing},
  author={Li, Runsheng and Ma, Hui and Wang, Rui and Song, Hao and Zhou, Xiangman and Wang, Lu and Zhang, Haiou and Zeng, Kui and Xia, Chunyang},
  journal={Journal of Manufacturing Processes},
  volume={143},
  pages={37--55},
  year={2025},
  publisher={Elsevier}
}

@article{deshpande2024deep,
  title={Deep learning-based image segmentation for defect detection in additive manufacturing: an overview},
  author={Deshpande, Sourabh and Venugopal, Vysakh and Kumar, Manish and Anand, Sam},
  journal={The International Journal of Advanced Manufacturing Technology},
  volume={134},
  number={5},
  pages={2081--2105},
  year={2024},
  publisher={Springer}
}

@article{yu2021fastflow,
  title={Fastflow: Unsupervised anomaly detection and localization via 2d normalizing flows},
  author={Yu, Jiawei and Zheng, Ye and Wang, Xiang and Li, Wei and Wu, Yushuang and Zhao, Rui and Wu, Liwei},
  journal={arXiv preprint arXiv:2111.07677},
  year={2021}
}

@article{jin2024pyramidal,
  title={Pyramidal flow matching for efficient video generative modeling},
  author={Jin, Yang and Sun, Zhicheng and Li, Ningyuan and Xu, Kun and Jiang, Hao and Zhuang, Nan and Huang, Quzhe and Song, Yang and Mu, Yadong and Lin, Zhouchen},
  journal={arXiv preprint arXiv:2410.05954},
  year={2024}
}

@article{dosovitskiy2020image,
  title={An image is worth 16x16 words: Transformers for image recognition at scale},
  author={Dosovitskiy, Alexey and Beyer, Lucas and Kolesnikov, Alexander and Weissenborn, Dirk and Zhai, Xiaohua and Unterthiner, Thomas and Dehghani, Mostafa and Minderer, Matthias and Heigold, Georg and Gelly, Sylvain and others},
  journal={arXiv preprint arXiv:2010.11929},
  year={2020}
}

@inproceedings{mishra2021vt,
  title={VT-ADL: A vision transformer network for image anomaly detection and localization},
  author={Mishra, Pankaj and Verk, Riccardo and Fornasier, Daniele and Piciarelli, Claudio and Foresti, Gian Luca},
  booktitle={2021 IEEE 30th International Symposium on Industrial Electronics (ISIE)},
  pages={01--06},
  year={2021},
  organization={IEEE}
}

@inproceedings{singh2023eval,
  title={Eval: Explainable video anomaly localization},
  author={Singh, Ashish and Jones, Michael J and Learned-Miller, Erik G},
  booktitle={Proceedings of the IEEE/CVF Conference on Computer Vision and Pattern Recognition},
  pages={18717--18726},
  year={2023}
}

@inproceedings{strater2024generalad,
  title={Generalad: Anomaly detection across domains by attending to distorted features},
  author={Str{\"a}ter, Luc PJ and Salehi, Mohammadreza and Gavves, Efstratios and Snoek, Cees GM and Asano, Yuki M},
  booktitle={European Conference on Computer Vision},
  pages={448--465},
  year={2024},
  organization={Springer}
}

@inproceedings{chen2025clep,
  title={CLEP: A Novel Contrastive Learning Method for Evolutionary Reentrancy Vulnerability Detection},
  author={Chen, Jie and Wang, Liangmin and Zhu, Huijuan and Sheng, Victor S},
  booktitle={Proceedings of the AAAI Conference on Artificial Intelligence},
  volume={39},
  number={1},
  pages={67--74},
  year={2025}
}

@article{he2025dualad,
  title={DualAD: exploring coupled dual-branch networks for multi-class unsupervised anomaly detection},
  author={He, Shiwen and Chen, Yuehan and Wang, Liangpeng and Huang, Wei and Xu, Rong and Qian, Yurong},
  journal={Electronics},
  volume={14},
  number={3},
  pages={594},
  year={2025},
  publisher={MDPI AG}
}

@inproceedings{yao2023one,
  title={One-for-all: Proposal masked cross-class anomaly detection},
  author={Yao, Xincheng and Zhang, Chongyang and Li, Ruoqi and Sun, Jun and Liu, Zhenyu},
  booktitle={Proceedings of the AAAI Conference on Artificial Intelligence},
  volume={37},
  number={4},
  pages={4792--4800},
  year={2023}
}

@inproceedings{tan2019encoder,
  title={An encoder-decoder based approach for anomaly detection with application in additive manufacturing},
  author={Tan, Yingshui and Jin, Baihong and Nettekoven, Alexander and Chen, Yuxin and Yue, Yisong and Topcu, Ufuk and Sangiovanni-Vincentelli, Alberto},
  booktitle={2019 18th IEEE international conference on machine learning and applications (ICMLA)},
  pages={1008--1015},
  year={2019},
  organization={IEEE}
}

@inproceedings{zhang2024multi,
  title={Multi-scale video anomaly detection by multi-grained spatio-temporal representation learning},
  author={Zhang, Menghao and Wang, Jingyu and Qi, Qi and Sun, Haifeng and Zhuang, Zirui and Ren, Pengfei and Ma, Ruilong and Liao, Jianxin},
  booktitle={Proceedings of the IEEE/CVF Conference on Computer Vision and Pattern Recognition},
  pages={17385--17394},
  year={2024}
}

@article{ bergmann2018improving,
  title={Improving unsupervised defect segmentation by applying structural similarity to autoencoders},
  author={Bergmann, Paul and L{\"o}we, Sindy and Fauser, Michael and Sattlegger, David and Steger, Carsten},
  journal={arXiv preprint arXiv:1807.02011},
  year={2018}
}

@inproceedings{gudovskiy2022cflow,
  title={Cflow-ad: Real-time unsupervised anomaly detection with localization via conditional normalizing flows},
  author={Gudovskiy, Denis and Ishizaka, Shun and Kozuka, Kazuki},
  booktitle={Proceedings of the IEEE/CVF winter conference on applications of computer vision},
  pages={98--107},
  year={2022}
}

@article{matsubara2020deep,
  title={Deep generative model using unregularized score for anomaly detection with heterogeneous complexity},
  author={Matsubara, Takashi and Sato, Kazuki and Hama, Kenta and Tachibana, Ryosuke and Uehara, Kuniaki},
  journal={IEEE Transactions on Cybernetics},
  volume={52},
  number={6},
  pages={5161--5173},
  year={2020},
  publisher={IEEE}
}

@article{livernoche2023diffusion,
  title={On diffusion modeling for anomaly detection},
  author={Livernoche, Victor and Jain, Vineet and Hezaveh, Yashar and Ravanbakhsh, Siamak},
  journal={arXiv preprint arXiv:2305.18593},
  year={2023}
}

@inproceedings{rudolph2022fully,
  title={Fully convolutional cross-scale-flows for image-based defect detection},
  author={Rudolph, Marco and Wehrbein, Tom and Rosenhahn, Bodo and Wandt, Bastian},
  booktitle={Proceedings of the IEEE/CVF winter conference on applications of computer vision},
  pages={1088--1097},
  year={2022}
}

@inproceedings{zavrtanik2021draem,
  title={Draem-a discriminatively trained reconstruction embedding for surface anomaly detection},
  author={Zavrtanik, Vitjan and Kristan, Matej and Sko{\v{c}}aj, Danijel},
  booktitle={Proceedings of the IEEE/CVF international conference on computer vision},
  pages={8330--8339},
  year={2021}
}

@inproceedings{venkataramanan2020attention,
  title={Attention guided anomaly localization in images},
  author={Venkataramanan, Shashanka and Peng, Kuan-Chuan and Singh, Rajat Vikram and Mahalanobis, Abhijit},
  booktitle={European Conference on Computer Vision},
  pages={485--503},
  year={2020},
  organization={Springer}
}

@article{xu2023unsupervised,
  title={Unsupervised industrial anomaly detection with diffusion models},
  author={Xu, Haohao and Xu, Shuchang and Yang, Wenzhen},
  journal={Journal of Visual Communication and Image Representation},
  volume={97},
  pages={103983},
  year={2023},
  publisher={Elsevier}
}

@inproceedings{ bergmann2020uninformed,
  title={Uninformed students: Student-teacher anomaly detection with discriminative latent embeddings},
  author={Bergmann, Paul and Fauser, Michael and Sattlegger, David and Steger, Carsten},
  booktitle={Proceedings of the IEEE/CVF conference on computer vision and pattern recognition},
  pages={4183--4192},
  year={2020}
}

@inproceedings{roth2022towards,
  title={Towards total recall in industrial anomaly detection},
  author={Roth, Karsten and Pemula, Latha and Zepeda, Joaquin and Sch{\"o}lkopf, Bernhard and Brox, Thomas and Gehler, Peter},
  booktitle={Proceedings of the IEEE/CVF conference on computer vision and pattern recognition},
  pages={14318--14328},
  year={2022}
}

@inproceedings{yi2020patch,
  title={Patch svdd: Patch-level svdd for anomaly detection and segmentation},
  author={Yi, Jihun and Yoon, Sungroh},
  booktitle={Proceedings of the Asian conference on computer vision},
  year={2020}
}

@inproceedings{liu2018future,
  title={Future frame prediction for anomaly detection--a new baseline},
  author={Liu, Wen and Luo, Weixin and Lian, Dongze and Gao, Shenghua},
  booktitle={Proceedings of the IEEE conference on computer vision and pattern recognition},
  pages={6536--6545},
  year={2018}
}

@misc{creality,
  title        = "Creality 3D Printers",
  howpublished = "\url{https://www.creality.com/}",
  year         = 2025,
  note         = "Accessed: 2025-01-08"
}

@article{haleem2019additive,
  title={Additive manufacturing applications in industry 4.0: a review},
  author={Haleem, Abid and Javaid, Mohd},
  journal={Journal of Industrial Integration and Management},
  volume={4},
  number={04},
  pages={1930001},
  year={2019},
  publisher={World Scientific}
}

@article{zhang2025diffusionad,
  title={DiffusionAD: Norm-guided one-step denoising diffusion for anomaly detection},
  author={Zhang, Hui and Wang, Zheng and Zeng, Dan and Wu, Zuxuan and Jiang, Yu-Gang},
  journal={IEEE transactions on pattern analysis and machine intelligence},
  year={2025},
  publisher={IEEE}
}

@inproceedings{luo2025exploring,
  title={Exploring intrinsic normal prototypes within a single image for universal anomaly detection},
  author={Luo, Wei and Cao, Yunkang and Yao, Haiming and Zhang, Xiaotian and Lou, Jianan and Cheng, Yuqi and Shen, Weiming and Yu, Wenyong},
  booktitle={Proceedings of the Computer Vision and Pattern Recognition Conference},
  pages={9974--9983},
  year={2025}
}

@inproceedings{bergmann2019mvtec,
  title={MVTec AD--A comprehensive real-world dataset for unsupervised anomaly detection},
  author={Bergmann, Paul and Fauser, Michael and Sattlegger, David and Steger, Carsten},
  booktitle={Proceedings of the IEEE/CVF conference on computer vision and pattern recognition},
  pages={9592--9600},
  year={2019}
}

@article{hasan2026engineering,
  title={Engineering Attack Vectors and Detecting Anomalies in Additive Manufacturing},
  author={Hasan, Md Mahbub and Sternhagen, Marcus and Roy, Krishna Chandra},
  journal={arXiv preprint arXiv:2601.00384},
  year={2026}
}
}


\end{document}